\documentclass[11pt]{article}
\usepackage[left=1in,right=1in,top=0.5in,bottom=0.5in,paper=a4paper]{geometry}
\usepackage{graphicx}
\usepackage{caption}

\title{\vspace{-40pt} Human Comprehensible Active Learning of Genome-Scale Metabolic Networks}

\author{
  \small\textbf{Lun Ai\textsuperscript{\rm 1}, Shi-Shun Liang\textsuperscript{\rm 2}, Wang-Zhou Dai\textsuperscript{\rm 3}, Liam Hallett\textsuperscript{\rm 2}, Stephen H. Muggleton\textsuperscript{\rm 1},} \\
  \small\textbf{Geoff S. Baldwin\textsuperscript{\rm 2}}\\ 
  \small\textsuperscript{\rm 1}Department of Computing, 
  \small\textsuperscript{\rm 2}Department of Life Science, Imperial College London, UK.\\
  \small\textsuperscript{\rm 3}School of Intelligence Science and Technology, Nanjing University, China. \\ 
  \small\{lun.ai15,shishun.liang20,l.hallett19, s.muggleton, g.baldwin\}@imperial.ac.uk, daiwz@nju.edu.cn
}

\date{}
\begin{document}

\maketitle

\noindent \textbf{Abtract} An important application of Synthetic Biology is the engineering of the host cell system to yield useful products. However, an increase in the scale of the host system leads to huge design space and requires a large number of validation trials with high experimental costs. A comprehensible machine learning approach that efficiently explores the hypothesis space and guides experimental design is urgently needed for reliable host cell engineering. We introduce a novel machine learning framework \textit{ILP-iML1515} based on Inductive Logic Programming (ILP). ILP-iML1515 actively learns abduced hypotheses of gene functions from auxotrophic mutant trials.  In contrast to numerical models, ILP-iML1515 is built on comprehensible logical representations of a genome-scale metabolic model of \textit{E.coli}, iML1515, and can update the model by learning new logical terms. We show that ILP-iML1515 framework 1) allows high-throughput simulations and 2) actively selects experiments that reduce the experimental cost of learning gene functions in comparison to randomly selected experiments. 

\section{Introduction}
\textit{Escherichia coli} (\textit{E.coli}) is a widely accepted recombinant host for synthesising key products. The most comprehensive metabolic model of E.coli \cite{iML1515} to date, iML1515, comprises 1515 genes and 2719 metabolic reactions. However, iML1515 is incomplete with respect to core gene functions and it has been demonstrated to contain errors \cite{bernstein_critical_2023}. Correction of the host metabolic network would improve its reliability, but the complexity of GEMs and experimental design space makes learning challenging both from the computational and experimental standpoint. Computational approaches to efficiently learn and navigate the design space would greatly enhance our ability to predictably engineer these biological systems.

We present preliminary results of a new framework ILP-iML1515 that uses Inductive Logic Programming (ILP) techniques \cite{ILP1991}. We show that active learning of abduced hypotheses can be applied to genome-scale metabolic network models (GEMs). This study examines the most comprehensive GEM of E. coli, iML1515 \cite{iML1515}. The biochemical, genetic, and genomic knowledge bases in iML1515 are compiled into logical representations that express the relationship between genes, reactions and metabolites. Such symbolic representations can be inspected by experts and are further processed into a novel matrix encoding for phenotype simulation. Through this new framework, we demonstrate a 4000-fold improvement in computational time, making whole-genome hypothesis generation and validation more accessible. We implement active learning to select auxotrophic mutant experiments that provide the maximal information entropy to discriminate candidate hypotheses with low experimental resource cost. To show the capability of ILP-iML1515, we delete gene functions in iML1515 and learn from hypotheses that cover all gene-enzyme associations in iML1515. The framework successfully recovers these removed gene functions from active learning with $ \frac{1}{10}$ of the experimental resource cost spent by randomly selected experiments.

\section{Related work}


Computational scientific discovery has been central to the research of Artificial Intelligence (AI). Modelling of biosynthetic systems typically involves genetic devices whose information can be effectively encoded in a symbolic and discrete format \cite{endy05}. Knowledge-driven AI utilises the domain knowledge to prune a discrete search space for symbolic hypotheses. In comparison, the quantitative engineering of biological pathways depends on numerical structures or constraints. This approach is represented by evolutionary and deep learning algorithms \cite{schmidt2009distilling,Udrescueaay2631,Neural_ODE} which learn numerical functions based on a large number of annotated data and extensive trial-and-errors. Owing to the lack of expressiveness, changing the structure of these machine learning systems and updating background knowledge are often difficult. The application of these systems in an experimental workflow is thus limited by flexibility, data efficiency and interpretability. 

Inductive Logic Programming (ILP) \cite{ILP1991} is a logic-based machine learning approach driven by both background knowledge and data. Examples, hypotheses, and background knowledge are represented by logic programs. Hypothesis generation and abduction which are core to scientific discovery, are facilitated by ILP \cite{TCIE}. Logic programs are interpretable representations and are well-suited for operating on structured knowledge bases such as biological models. The Robot Scientist \cite{King04:RobotScientist} demonstrated that abductive learning could be used to learn metabolic networks faster and at less cost than random experiments. However, this demonstration was based on a metabolic network of only 17 genes in aromatic amino acid pathways. In contrast, our ILP-iML1515 framework implements the genome-scale metabolic network model iML1515 that includes 1515 genes and 2719 reactions. 

\section{Framework}

\begin{figure}[h]
\begin{minipage}{.30\textwidth}
    \centering 
  \begin{tabular}{c|c}
		  & No. Genes \\
		\hline
             & \\
		Robot & 17 \\ 
            Scientist & \\
            & \\\hline
            & \\
		ILP-iML1515 & 1515 \\
            & \\
	\end{tabular}%
\captionof{table}{Model sizes.}
\label{table:model_size}
\end{minipage}
\hfill
\begin{minipage}{.67\textwidth}
\centering
\includegraphics[width=\textwidth]{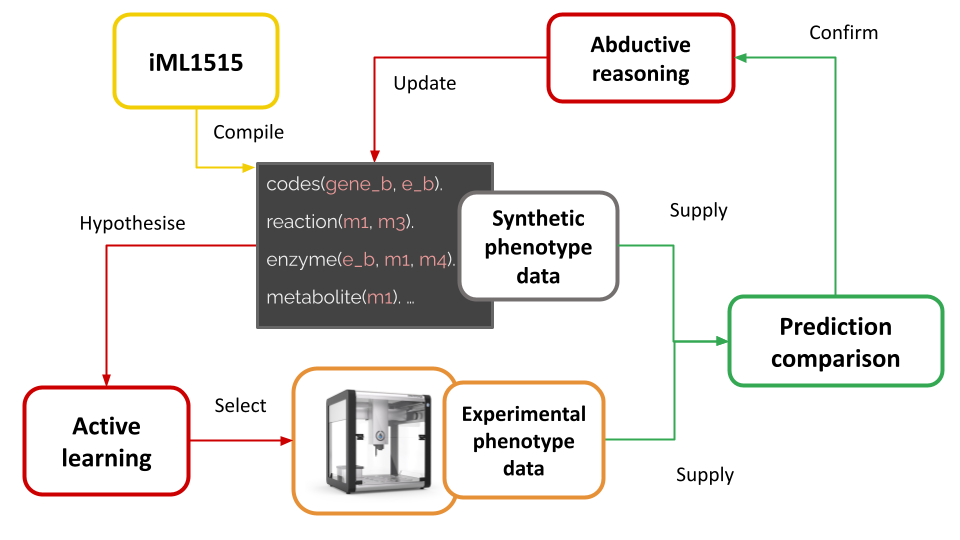}
\captionof{figure}{ILP-iML1515 framework.}
\label{fig:framework}
\end{minipage}
\end{figure}

ILP-iML1515 operates on a model that is nearly 90 times larger than the model encoded in the Robot Scientist (Table \ref{table:model_size}). Our framework (Figure \ref{fig:framework}) takes discretised data that describe phenotype observations in selected auxotrophic mutant experiments. The outcomes of the experiments are supplied by online databases or the laboratory.    

We implement a novel matrix encoding of metabolic reactions in iML1515. iML1515 is first converted into logical representations that express the coding function of genes ($codes$, $enzyme$) and the involvement of metabolites in reactions ($reaction$, $metabolites$). Figure \ref{fig:framework} shows an example of these logical terms. Based on these terms, we assemble several binary vectors and matrices. These are used to saturate the synthesisable set of metabolites by mutant strains. Metabolite saturation is dependent on the knockout gene of the auxotrophic mutants and the growth media. 

Given a defined set of hypothesised gene functions, synthetic phenotypic data are simulated by applying binary operations to binary vectors and matrices. Binary operations compute the closure set of synthesisable metabolites and we make a binary phenotypic classification according to the inclusion of essential metabolites. Simulated phenotypes are assessed against experimental observations. 

Test outcomes are supplied to the abductive reasoning module where we confirm hypotheses consistent with the experimental outcomes and prune the inconsistent hypotheses. All candidate hypotheses are generated from compiled knowledge bases. The remaining hypotheses are tested with further experiments suggested by active learning. Based on information entropy, the active learning module selects experiments that maximise the discrimination between candidate hypotheses and minimise the expected cost of experimental resources. To complete the cycle, experiments that are actively selected will be conducted in the laboratory or consulted from online data sources. 


\section{Results}
\begin{table}[t]
\centering
	\begin{tabular}{c|c|c}
		Wall time per simulation (sec) & Robot Scientist & ILP-iML1515 \\ & & \\
		\hline
		& & \\
		No parallelisation & 250 & 0.6\\ 
		& & \\ \hline
		& & \\
		20 CPUs & 27 & 0.06 \\ 
	\end{tabular}
\bigskip
\caption{Simulation run-time efficiency.}
\label{table:runtime_improvement}
\end{table}

\begin{figure}[h]
\centering
\includegraphics[width=0.9\textwidth]{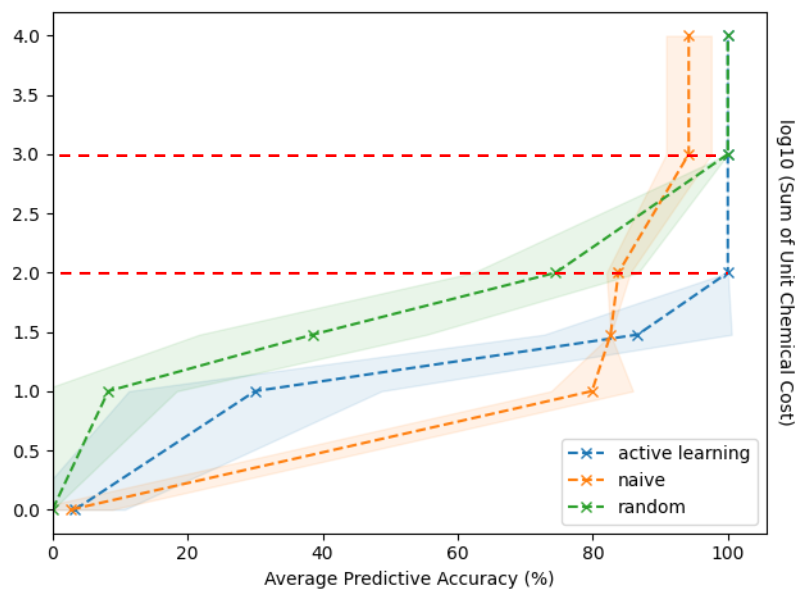}
\caption{Experimental cost reduction. }
\label{fig:cost_reduction}
\end{figure}

In GEMs, abduction of gene function hypotheses and in-vivo validations requires exploring large hypothesis spaces and design spaces. Both are challenging due to the complexity of these models and the cost of conducting physical experiments. Our results show that the ILP-iML1515 framework can efficiently perform phenotype simulations and actively suggest informative experiments to be performed in the laboratory. Based on abductive logical reasoning \cite{TCIE} and active learning \cite{King04:RobotScientist}, we automate the selection of experiments to optimise experimental resources. 

In Table \ref{table:runtime_improvement}, we examine the computational efficiency of our new logical matrix encoding against the metabolic model encoding in the Robot Scientist \cite{King04:RobotScientist}. The Robot Scientist encoding of iML1515 has been adapted to the same logic programming environment (SWI-Prolog) as ILP-iML1515. We compare the average simulation time of our new logical matrix encoding against the metabolic model encoding in Robot Scientist. We take advantage of the parallelisation with this encoding and computation runtime has been improved significantly. With multiple core parallelisation, logical matrix encoding uses $\frac{1}{4000}$ of the simulation time. 

In Figure \ref{fig:cost_reduction}, we compare the effect of actively selected experiments for recovering known gene functions.  ILP-iML1515 was used to generate synthetic data of 23617 single-knockout trials \cite{iML1515}. Active learning algorithm (ase) is contrasted against two alternative trial selection methods: selection of the cheapest trials (naive) and random trial selection (random). We randomly sampled three genes in the iML1515 model. To create the incomplete models, we removed one gene function from the three. For abduction, we generated potential gene-enzyme mappings for all enzyme functions in iML1515. We recover the model by introducing the consistent abduced hypothesis. The synthetic trial outcomes are compared against trial outcomes predicted by recovered models. The y-axis is the sum of the chemical cost in $log_{10}$ when selected trials are performed. Since predictions of recovered models are compared with synthetic data, 100\% predictive accuracy indicates successful recovery of the deleted gene functions. Both active learning (blue) and random trial selection (green) enable the correct gene functions to be learned (100\% accuracy). However, active learning only uses 1/10 the cost of random selection (red dotted lines).



\section{Future Work}

Future work aims to integrate this approach with a high-throughput experimental workflow. The focus will be to use CRISPRi, an advanced biological gene editing tool, to target repress genes for experimental confirmation of candidate hypotheses during active learning. Importantly this approach will enable us to address multiple gene-loci in combination, which are not addressable by current mutagenesis approaches.

\vspace{.5em}
\large
\bibliographystyle{abbrv}
\bibliography{ilp_iml1515}

\begin{thebibliography}{1}

\bibitem{bernstein_critical_2023}
D.~B. Bernstein, B.~Akkas, M.~N. Price, and A.~P. Arkin.
\newblock Critical assessment of {E}. coli genome-scale metabolic model with
  high-throughput mutant fitness data, 2023.

\bibitem{Neural_ODE}
R.~T.~Q. Chen, Y.~Rubanova, J.~Bettencourt, and D.~Duvenaud.
\newblock Neural ordinary differential equations.
\newblock In {\em Proceedings of the 32nd International Conference on Neural
  Information Processing Systems}, page 6572–6583, 2018.

\bibitem{endy05}
D.~Endy.
\newblock {Foundations for Engineering Biology}.
\newblock {\em Nature}, 438:449--53, 12 2005.

\bibitem{King04:RobotScientist}
R.~D. King, K.~E. Whelan, F.~M. Jones, P.~G.~K. Reiser, C.~H. Bryant, S.~H.
  Muggleton, D.~B. Kell, and S.~G. Oliver.
\newblock Functional genomic hypothesis generation and experimentation by a
  robot scientist.
\newblock {\em Nature}, 427:247--252, 2004.

\bibitem{iML1515}
J.~M. Monk, C.~J. Lloyd, E.~Brunk, N.~Mih, A.~Sastry, Z.~King, R.~Takeuchi,
  W.~Nomura, Z.~Zhang, H.~Mori, A.~M. Feist, and B.~O. Palsson.
\newblock {iML1515,} a knowledgebase that computes escherichia coli traits.
\newblock {\em Nature Biotechnology}, 35(10):904–908, 2017.

\bibitem{ILP1991}
S.~H. Muggleton.
\newblock Inductive logic programming.
\newblock {\em New Generation Computing}, 8:295–318, 1991.

\bibitem{TCIE}
S.~H. Muggleton and C.~H. Bryant.
\newblock Theory completion using inverse entailment.
\newblock In {\em Proceedings of the 10th International Workshop on Inductive
  Logic Programming (ILP-00)}, pages 130--146, 2000.

\bibitem{schmidt2009distilling}
M.~Schmidt and H.~Lipson.
\newblock Distilling free-form natural laws from experimental data.
\newblock {\em science}, 324(5923):81--85, 2009.

\bibitem{Udrescueaay2631}
S.~Udrescu and M.~Tegmark.
\newblock {AI Feynman}: A physics-inspired method for symbolic regression.
\newblock {\em Science Advances}, 6(16), 2020.

\end{thebibliography}

\end{document}